\documentclass[11pt,a4paper]{article}
\usepackage[hyperref]{acl2021}
\usepackage{times}
\usepackage{latexsym}

\usepackage{microtype}

\aclfinalcopy 
\def\aclpaperid{1785} 

\usepackage{booktabs}
\newcommand{\tabincell}[2]{\begin{tabular}{@{}#1@{}}#2\end{tabular}}
\usepackage{graphicx}
\usepackage{subfig}
\usepackage{bm}
\usepackage{amsmath}
\usepackage{CJKutf8}

\title{MECT: Multi-Metadata Embedding based Cross-Transformer for Chinese Named Entity Recognition}

\author{Shuang Wu$^1$, ~Xiaoning Song$^{1*}$, ~Zhenhua Feng$^{2, 3}$\thanks{~~Corresponding author.}\\
  $^1$School of Artificial Intelligence and Computer Science, Jiangnan University, China\\
  $^2$Department of Computer Science, University of Surrey, UK\\
  $^3$Centre for Vision, Speech and Signal Processing, University of Surrey, UK\\
  \texttt{shuangwu@stu.jiangnan.edu.cn}
  \\
  \texttt{x.song@jiangnan.edu.cn, z.feng@surrey.ac.uk}
  }

\date{}

\begin{document}
\begin{CJK}{UTF8}{gbsn}
\maketitle
\begin{abstract}
Recently, word enhancement has become very popular for Chinese Named Entity Recognition (NER), reducing segmentation errors and increasing the semantic and boundary information of Chinese words. However, these methods tend to ignore the information of the Chinese character structure after integrating the lexical information. Chinese characters have evolved from pictographs since ancient times, and their structure often reflects more information about the characters. This paper presents a novel Multi-metadata Embedding based Cross-Transformer (MECT) to improve the performance of Chinese NER by fusing the structural information of Chinese characters. Specifically, we use multi-metadata embedding in a two-stream Transformer to integrate Chinese character features with the radical-level embedding. With the structural characteristics of Chinese characters, MECT can better capture the semantic information of Chinese characters for NER. The experimental results obtained on several well-known benchmarking datasets demonstrate the merits and superiority of the proposed MECT method.\footnote{The source code of the proposed method is publicly available at \url{https://github.com/CoderMusou/MECT4CNER}.} 
\end{abstract}

\section{Introduction}
Named Entity Recognition (NER) plays an essential role in structuring of unstructured text. It is a sequence tagging task that extracts named entities from unstructured text. Common categories of NER include names of people, places, organizations, time, quantity, currency, and some proper nouns. NER is the basis for many Natural Language Processing (NLP) tasks such as event extraction~\cite{chen2015event}, question answering~\cite{diefenbach2018core}, information retrieval~\cite{khalid2008impact},  knowledge graph construction~\cite{riedel2013relation}, etc.

\begin{table}[!t]
\centering
\small \setlength{\tabcolsep}{12.5pt}
\begin{tabular}{lccc}
\specialrule{0.1em}{3pt}{3pt}
\textbf{Character} & \textbf{CR} & \textbf{HT} & \textbf{SC} \\
\specialrule{0.1em}{3pt}{3pt}
\specialrule{0em}{1pt}{1pt}
题 (topic) & 页 & 是页 & 日一走页 \\
\specialrule{0em}{1pt}{1pt}
榆 (elm)   & 木 & 木俞 & 木人一月刂 \\
\specialrule{0em}{1pt}{1pt}
渡 (ferry) & 氵 & 氵度 & 氵广廿又 \\
\specialrule{0em}{1pt}{1pt}
脸 (face)  & 月 & 月佥 & 月人一ツ一 \\
\specialrule{0.1em}{3pt}{3pt}
\end{tabular}
\caption{\label{Chinese-structure} Structure decomposition of Chinese characters: `CR' denotes the Chinese radical, `HT' denotes the head and tail, and `SC' denotes the structural components of Chinese characters. }
\end{table}

\begin{table*}
\small
\centering 
\setlength{\tabcolsep}{11pt}
\begin{tabular}{ccl}
\specialrule{0.1em}{3pt}{3pt}
\textbf{Radicals} & \textbf{Denotation} & \textbf{Examples} \\
\specialrule{0.1em}{3pt}{3pt}
\specialrule{0em}{1pt}{1pt}
鸟 (\texttt{\small bird}) & birds & 鸡 (\texttt{\small chicken}),\quad 鸭 (\texttt{\small duck}),\quad 鹅 (\texttt{\small goose}),\quad 鹰 (\texttt{\small eagle}) \\
\specialrule{0em}{1pt}{1pt}
艹 (\texttt{\small grass}) & herbaceous plants & 花 (\texttt{\small flower}),\quad 草 (\texttt{\small grass}),\quad 菜 (\texttt{\small vegetable}),\quad 茶 (\texttt{\small tea}) \\
\specialrule{0em}{1pt}{1pt}
月 (\texttt{\small meat}) & body parts & 肾 (\texttt{\small kidney}),\quad 脚 (\texttt{\small foot}),\quad 腿 (\texttt{\small leg}),\quad 脑 (\texttt{\small brain}) \\
\specialrule{0em}{1pt}{1pt}
\specialrule{0.1em}{3pt}{3pt}
\end{tabular}
\caption{\label{radicals-meaning} Some examples of Chinese radicals, including `鸟'(\texttt{\small bird}), `艹'(\texttt{\small grass}) and `月'(\texttt{\small meat}).}
\end{table*}

Compared with English, there is no space between Chinese characters as word delimiters. Chinese word segmentation is mostly distinguished by readers through the semantic information of sentences, posing many difficulties to Chinese NER~\cite{duan2011study, ma-etal-2020-simplify}. Besides, the task also has many other challenges, such as complex combinations, entity nesting, and indefinite length~\cite{dong2016character}.

In English, different words may have the same root or affix that better represents the word's semantics. For example, physiology, psychology, sociology, technology and zoology contain the same suffix, `-logy', which helps identify the entity of a subject name. Besides, according to the information of English words, root or affixes often determine general meanings~\citep{yadav-etal-2018-deep}. The root, such as `ophthalmo-' (\texttt{\small ophthalmology}), `esophage-' (\texttt{\small esophagus}) and `epithelio-' (\texttt{\small epithelium}), can help human or machine to better recognize professional nouns in medicine. Therefore, even the state-of-the-art methods, such as BERT~\cite{devlin-etal-2019-bert} and GPT~\cite{radford2018improving}, trained on large-scale datasets, adopt this delicate word segmentation method for performance boost.

For Chinese characters, there is also a structure similar to the root and affixes in English. According to the examples in Table \ref{Chinese-structure}, we can see that the structure of Chinese characters has different decomposition methods, including the Chinese radical (CR), head and tail (HT) and structural components (SC). Chinese characters have evolved from hieroglyphs since ancient times, and their structure often reflects more information about them. There are some examples in Table \ref{radicals-meaning}. The glyph structure can enrich the semantics of Chinese characters and improve the performance of NER. 
For example, the Bi-LSTM-CRF method~\cite{dong2016character} firstly obtains character-level embedding through the disassembly of Chinese character structure to improve the performance of NER. However, LSTM is based on time series modeling, and the input of each cell depends on the output of the previous cell. So the LSTM-based model is relatively complicated and the parallel ability is limited.

To address the aforementioned issues, we take the advantages of Flat-Lattice Transformer (FLAT)~\cite{li-etal-2020-flat} in efficient parallel computing and excellent lexicon learning, and introduce the radical stream as an extension on its basis. By combining the radical information, we propose a Multi-metadata Embedding based Cross-Transformer (MECT). MECT has the lattice- and radical-streams, which not only possesses FLAT's word boundary and semantic learning ability but also increases the structure information of Chinese character radicals. This is very effective for NER tasks, and has improved the baseline method on different benchmarks. The main contributions of the proposed method include:
\begin{itemize}
    \item The use of multi-metadata feature embedding of Chinese characters in Chinese NER.
    
    \item A novel two-stream model that combines the radicals, characters and words of Chinese characters to improve the performance of the proposed MECT method.
    
    \item The proposed method is evaluated on several well-known Chinese NER benchmarking datasets, demonstrating the merits and superiority of the proposed approach over the state-of-the-art methods. 
\end{itemize}

\section{Related Work}
The key of the proposed MECT method is to use the radical information of Chinese characters to enhance the Chinese NER model. So we focus on the mainstream information enhancement methods in the literature. There are two main types of Chinese NER enhancement methods, including lexical information fusion and glyph-structural information fusion.

\textbf{Lexical Enhancement} In Chinese NER, many recent studies use word matching methods to enhance character-based models. A typical method is the Lattice-LSTM model~\cite{zhang-yang2018chinese} that improves the NER performance by encoding and matching words in the lexicon. Recently, some lexical enhancement methods were proposed using CNN models, such as LR-CNN~\cite{gui2019cnn}, CAN-NER~\cite{zhu-wang-2019-ner}. Graph networks have also been used with lexical enhancement. The typical one is LGN~\cite{gui2019lexicon}. Besides, there are Transformer-based lexical enhancement methods, such as PLT~\cite{xue2019porous} and FLAT. And SoftLexicon~\cite{ma-etal-2020-simplify} introduces lexical information through label and probability methods at the character representation layer.

\textbf{Glyph-structural Enhancement}  
Some studies also use the glyph structure information in Chinese NER. For example, \citet{dong2016character} were the first to study the application of radical-level information in Chinese NER.
They used Bi-LSTM to extract radical-level embedding and then concatenated it with the embedding of characters as the final input. The radical information used in Bi-LSTM is structural components (SC) as shown in Table \ref{Chinese-structure}, which achieved state-of-the-art performance on the MSRA dataset. The Glyce~\cite{meng2019glyce} model used Chinese character images to extract features such as strokes and structure of Chinese characters, achieving promising performance in Chinese NER. Some other methods~\cite{xu2019exploiting, song2020joint} also proposed to use radical information and Tencent's pre-trained embedding\footnote{\url{https://ai.tencent.com/ailab/nlp/en/embedding.html}} to improve the performance.
In these works, the structural components of Chinese characters have been proven to be able to enrich the semantics of the characters, resulting in better NER performance. 

\section{Background}
The proposed method is based on the Flat-Lattice Transformer (FLAT) model. Thus, we first briefly introduce FLAT that improves the encoder structure of Transformer by adding word lattice information, including semantic and position boundary information. These word lattices are obtained through dictionary matching. 

\begin{figure}[!t]
\centering
\includegraphics[scale=0.26]{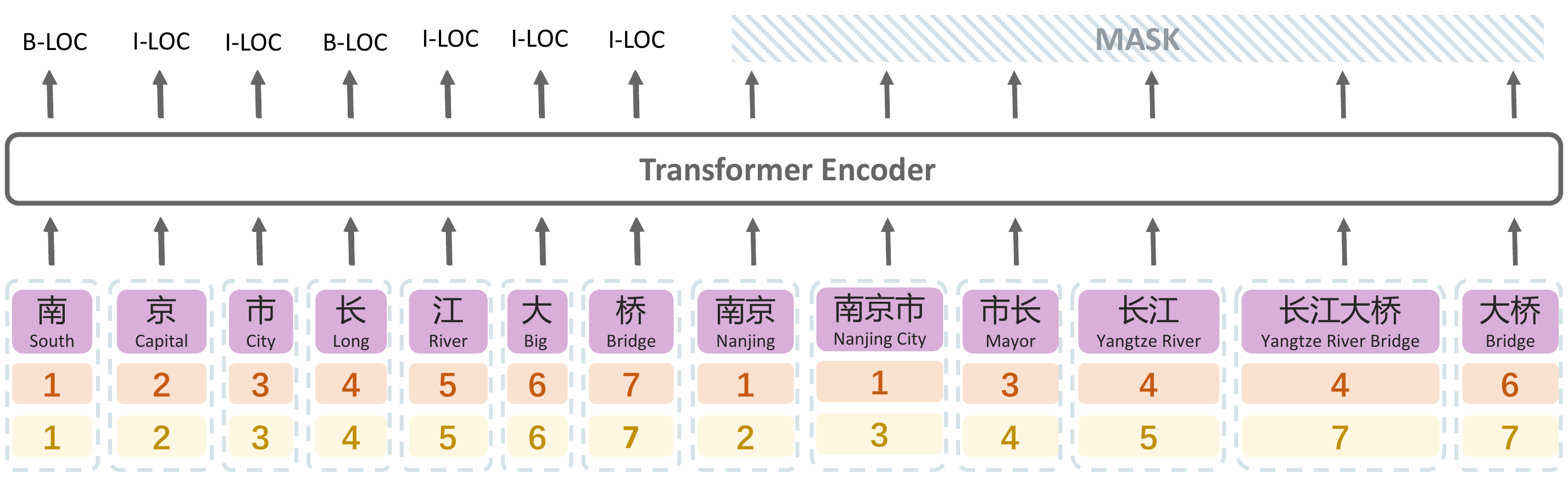}
\caption{\label{FLAT-in-out}The input and output of FLAT.}
\end{figure}

Figure \ref{FLAT-in-out} shows the input and output of FLAT. It uses the relative position encoding transformed by head and tail position to fit the word's boundary information. The relative position encoding, $\bm{R}_{ij}$, is calculated as follows:
\begin{equation}
\begin{split}
  \bm{R}_{ij} &= \text{ReLU}(\bm{W}_r(\bm{p}_{h_i - h_j} \oplus \bm{p}_{h_i - t_j} \\& \oplus \bm{p}_{t_i - h_j}  \oplus \bm{p}_{t_i - t_j})),\label{R_ij}
\end{split}
\end{equation}
where $\bm{W}_r$ is a learnable parameter, $h_i$ and $t_i$ represent the head position and tail position of the i-th character, $\oplus$ denotes the concatenation operation, and $\bm{p}_{span}$ is obtained as in ~\citet{vaswani2017attention}:
\begin{align}
  \bm{p}_{span}^{(2k)} &= \sin(\frac{span}{10000^{2k/d_{model}}}),\\
  \bm{p}_{span}^{(2k+1)} &= \cos(\frac{span}{10000^{2k/d_{model}}}),
\end{align}
where $\bm{p}_{span}$ corresponds to $\bm{p}$ in Eq.~(\ref{R_ij}), and $span$ denotes $h_i - h_j$, $h_i - t_j$, $t_i - h_j$ and $t_i - t_j$. Then the scaled dot-product attention is obtained by:
\begin{gather}
  Att(\bm{A}, \bm{V}) = \text{softmax}(\bm{A})\bm{V},\\
  \bm{A}_{ij} = (\bm{Q}_i + \bm{u})^\top\bm{K}_j + (\bm{Q}_i + \bm{v})^\top\bm{R}^*_{ij},\\
  [\bm{Q}, \bm{K}, \bm{V}] = E_x[\bm{W}_q, \bm{W}_k, \bm{W}_v],
\end{gather}
where $\bm{R}^*_{ij} = \bm{R}_{ij}\cdot\bm{W}_{R}$. $\bm{u}$, $\bm{v}$ and $\bm{W}_\Box$ are learnable parameters.

\begin{figure*}[!t]  
    \centering    
    \subfloat[\label{whole-frame}The whole architecture] 
    {
        \begin{minipage}[t]{0.62\textwidth}
            \centering          
            \includegraphics[width=1\textwidth]{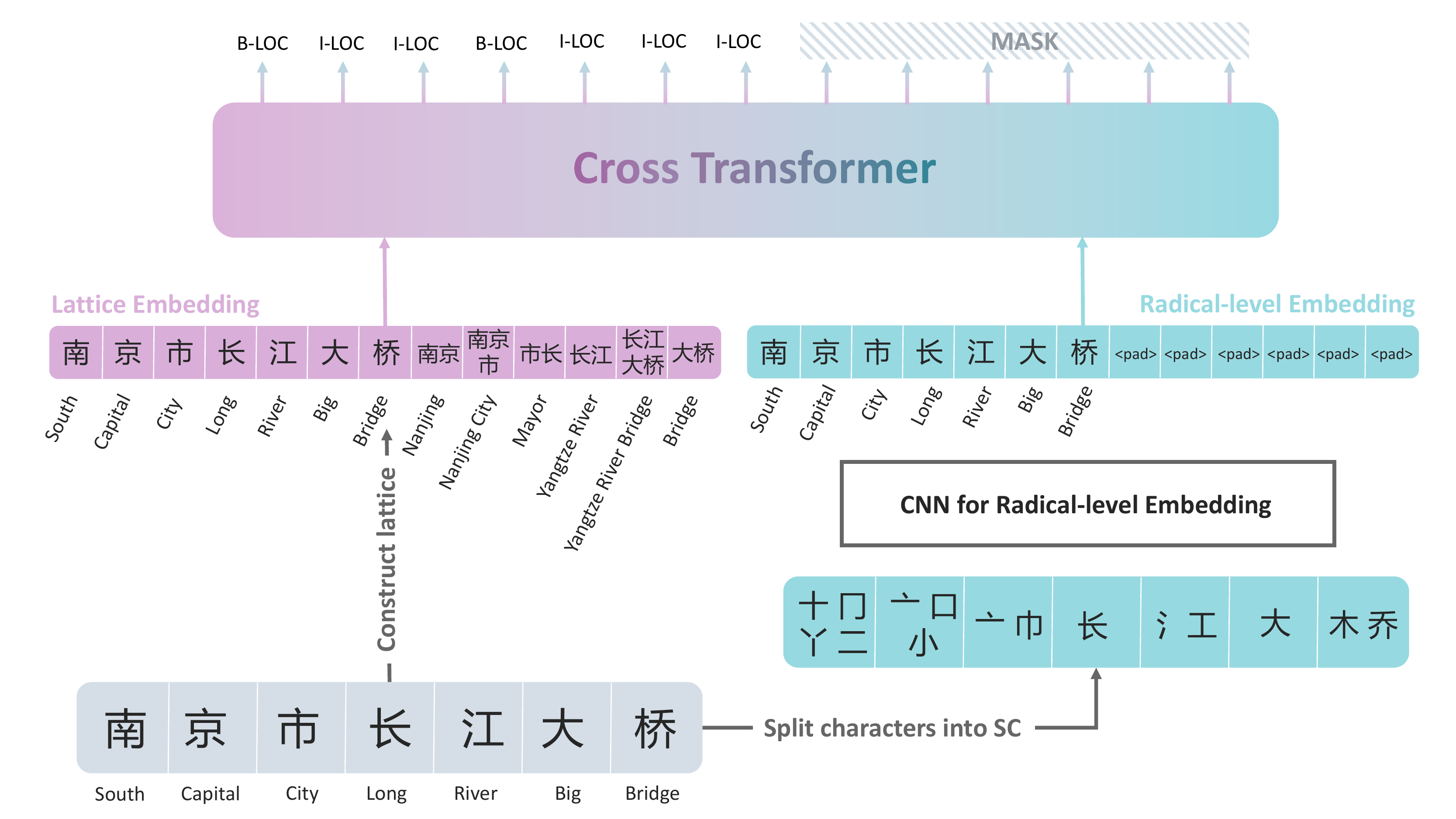}   
        \end{minipage}%
    }
    \subfloat[\label{cross-transformer}The Cross-Transformer module] 
    {
        \begin{minipage}[t]{0.38\textwidth}
            \centering      
            \includegraphics[width=1\textwidth]{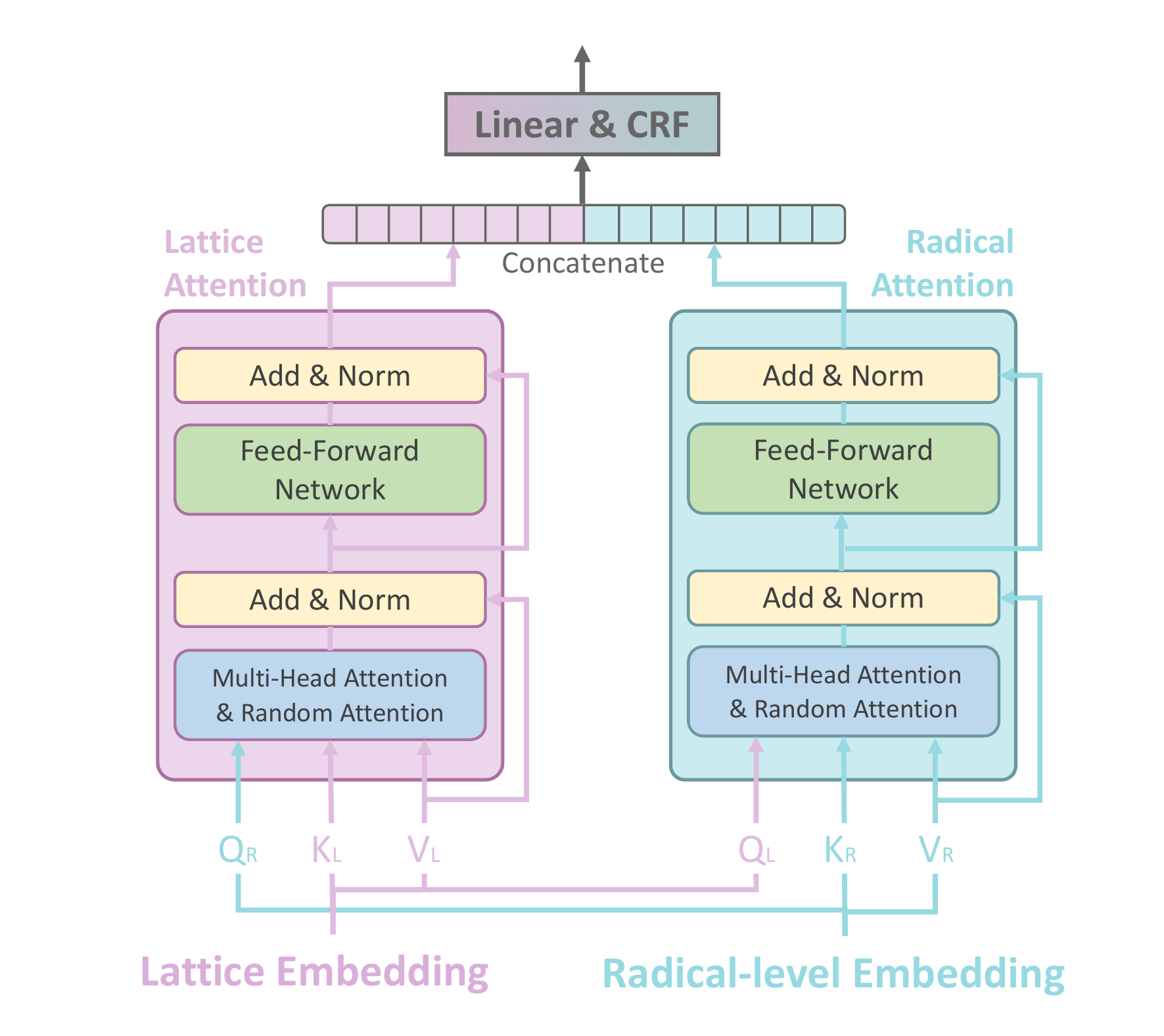}   
        \end{minipage}
    }
    \caption{\label{model-structure} The proposed MECT method: (a) the overall structure of MECT; (b) the Cross-Transformer module. } 
\end{figure*}

\section{The Proposed MECT Method}
To better integrate the information of Chinese character components, we use Chinese character structure as another metadata and design a two-stream form of multi-metadata embedding network. The architecture of the proposed network is shown in Figure \ref{whole-frame}. The proposed method is based on the encoder structure of Transformer and the FLAT method, in which we integrate the meaning and boundary information of Chinese words. The proposed two-stream model uses a Cross-Transformer module similar to the self-attention structure to fuse the information of Chinese character components. In our method, we also use the multi-modal collaborative attention method that is widely used in vision-language tasks~\cite{lu2019vibert}. The difference is that we add a randomly initialized attention matrix to calculate the attention bias for the two types of metadata embedding.

\begin{figure}[!t]
\centering
\includegraphics[scale=0.4]{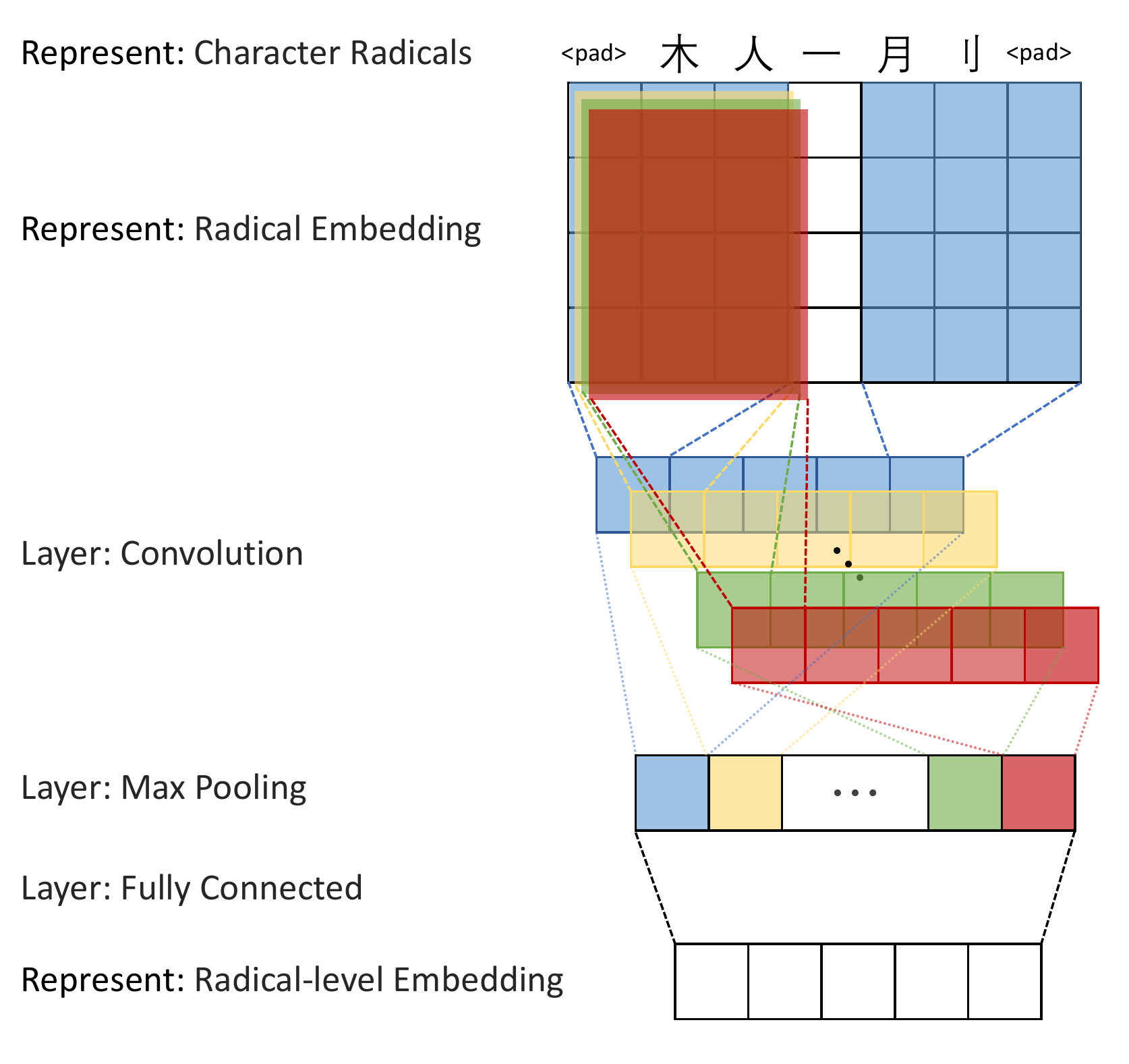}
\caption{\label{CNN-embed}CNN for radical feature extraction.}
\end{figure}

\subsection{CNN for Radical-level Embedding}
Chinese characters are based on pictographs, and their meanings are expressed in the shape of objects. In this case, the structure of Chinese characters has certain useful information for NER. For example, the radicals such as `艹' (\texttt{\small grass}) and `木' (\texttt{\small wood}) generally represent plants, enhancing Chinese medicine entity recognition. For another example, `月' (\texttt{\small body}) represents human body parts or organs, and `疒' (\texttt{\small disease}) represents diseases, which benefits Chinese NER for the medical field. Besides, the Chinese have their own culture and belief in naming. Radicals `钅' (\texttt{\small metal}), `木' (\texttt{\small wood}), `氵' (\texttt{\small water}), `火' (\texttt{\small fire}), and `土' (\texttt{\small earth}) represented by the Wu-Xing (Five Elements) theory are often used as names of people or companies. But `锈' (\texttt{\small rust}), `杀' (\texttt{\small kill}), `污' (\texttt{\small dirt}), `灾' (\texttt{\small disaster}) and `堕' (\texttt{\small fall}) are usually not used as names, even if they contain some elements of the Wu-Xing theory. It is because the other radical components also determine the semantics of Chinese characters. Radicals that generally appear negative or conflict with Chinese cultural beliefs are usually not used for naming.

Therefore, we choose the more informative Structural Components (SC) in Table~\ref{Chinese-structure} as radical-level features of Chinese characters and use Convolutional Neural Network (CNN) to extract character features. The structure diagram of the CNN network is shown in Figure \ref{CNN-embed}. We first disassemble the Chinese characters into SC and then input the radicals into CNN. Last, we use the max-pooling and fully connected layers to get the feature embedding of Chinese characters at the radical-level. 

\subsection{The Cross-Transformer Module}
After radical feature extraction, we propose a Cross-Transformer network to obtain the supplementary semantic information of the structure of Chinese characters. It also uses contextual and lexical information to enrich the semantics of Chinese characters. The Cross-Transformer network is illustrated in Figure \ref{cross-transformer}. We use two Transformer encoders to cross the lattice and radical information of Chinese characters, which is different from the self-attention method in Transformer.

The input $\bm{Q}_L(\bm{Q}_R), \bm{K}_L(\bm{K}_R), \bm{V}_L(\bm{V}_R)$ are obtained by the linear transformation of lattice and radical-level feature embedding:
\begin{gather}
  \begin{bmatrix}
  	\bm{Q}_{L(R),i}\\
  	\bm{K}_{L(R),i}\\
  	\bm{V}_{L(R),i}
  \end{bmatrix}^\top
   = \bm{E}_{L(R),i}
  \begin{bmatrix}
  	\bm{W}_{L(R),Q}\\
  	\bm{I}\\
  	\bm{W}_{L(R),V}
  \end{bmatrix}^\top,
\end{gather}
where $\bm{E}_{L}$ and $\bm{E}_{R}$ are lattice embedding and radical-level embedding, $\bm{I}$ is the identity matrix, and each $\bm{W}$ is a learnable parameter. Then we use the relative position encoding in FLAT to represent the boundary information of a word and calculate the attention score in our Cross-Transformer:
\begin{eqnarray}
\text{Att}_L(\bm{A}_R, \bm{V}_L)=\text{Softmax}(\bm{A}_R)\bm{V}_L,\\
\text{Att}_R(\bm{A}_L, \bm{V}_R)=\text{Softmax}(\bm{A}_L)\bm{V}_R,\\
\begin{aligned}
  \bm{A}_{L(R), ij}  &= (\bm{Q}_{L(R),i} + \bm{u}_{L(R)})^\top\bm{K}_{R(L),j}\\
  				    &+ (\bm{Q}_{L(R),i} + \bm{v}_{L(R)})^\top\bm{R}^*_{L(R),ij},\label{attenion_matrix}
\end{aligned}
\end{eqnarray}
where $\bm{u}$ and $\bm{v}$ are learnable parameters for attention bias in Eq.~(\ref{attenion_matrix}), $\bm{A}_L$ is the lattice attention score, and $\bm{A}_R$ denotes the radical attention score. And $\bm{R}^*_{ij} = \bm{R}_{ij}\cdot\bm{W}_{R}$. $\bm{W}_R$ are learnable parameters. The relative position encoding, $\bm{R}_{ij}$, is calculated as follows:
\begin{equation}
\begin{split}
  \bm{R}_{ij} &= \text{ReLU}(\bm{W}_r(\bm{p}_{h_i - h_j} \oplus \bm{p}_{t_i - t_j})).
\end{split}
\end{equation}

\subsection{Random Attention}
We empirically found that the use of random attention in Cross-Transformer can improve the performance of the proposed method. This may be due to the requirement of attention bias in lattice and radical feature embedding, which can better adapt to the scores of two subspaces. Random attention is a randomly initialized parameter matrix $\bm{B}^{max\_len\times max\_len}$ that is added to the previous attention score to obtain a total attention score:
\begin{eqnarray}
\bm{V}^*_L = \text{Softmax}(\bm{A}_R+\bm{B})\bm{V}_L,\\
\bm{V}^*_R = \text{Softmax}(\bm{A}_L+\bm{B})\bm{V}_R.
\end{eqnarray}

\subsection{The Fusion Method}
To reduce information loss, we directly concatenate the lattice and radical features and input them into a fully connected layer for information fusion:
\begin{equation}
\text{Fusion}(\bm{V}^*_L, \bm{V}^*_R)=(\bm{V}^*_R \oplus \bm{V}^*_L) \bm{W}^o + \bm{b},
\end{equation}
where $\oplus$ denotes the concatenation operation, $\bm{W}^o$ and $\bm{b}$ are learnable parameters.

After the fusion step, we mask the word part and pass the fused feature to a Conditional Random Field (CRF)~\cite{Lafferty:2001:CRF:645530.655813} module.

\section{Experimental Results}
In this section, we evaluate the proposed MECT method on four datasets. To make the experimental results more reasonable, we also set up two additional working methods for assessing the performance of radicals in a two-stream model. We use the span method to calculate F1-score (F1), precision (P), and recall (R) as the evaluation metrics.

\subsection{Experimental Settings}
We use four mainstream Chinese NER benchmarking datasets: Weibo~\cite{peng-dredze-2015-named, HeS16}, Resume~\cite{zhang-yang2018chinese}, MSRA~\cite{levow-2006-third}, and Ontonotes 4.0~\cite{trove.nla.gov.au/work/192067053}. The corpus of MSRA and Ontonotes 4.0 comes from news, the corpus of Weibo comes from social media, and the corpus of Resume comes from the resume data in Sina Finance. Table \ref{statistics} shows the statistical information of these datasets. Among them, the Weibo dataset has four types of entities, including PER, ORG, LOC, and GPE. Resume has eight types of entities, including CONT, EDU, LOC, PER, ORG, PRO, RACE, and TITLE. OntoNotes 4.0 has four types of entities: PER, ORG, LOC, and GPE. The MSRA dataset contains three types of entities, \textit{i.e.}, ORG, PER, and LOC.
\begin{table}
\centering\small \setlength{\tabcolsep}{8pt}
\begin{tabular}{lcccc}
\specialrule{0.1em}{3pt}{3pt}
\textbf{Datasets} & \textbf{Types} & \textbf{Train} & \textbf{Dev} & \textbf{Test} \\
\specialrule{0.1em}{3pt}{3pt}
\specialrule{0em}{1pt}{1pt}
	Weibo 
	& \tabincell{c}{Sentences \\Entities} 
	& \tabincell{c}{1.35k \\1.89k}
	& \tabincell{c}{0.27k \\0.39k}
	& \tabincell{c}{0.27k \\0.42k} \\
\specialrule{0.05em}{1pt}{1pt}
	Resume 
	& \tabincell{c}{Sentences \\Entities} 
	& \tabincell{c}{3.8k \\1.34k}
	& \tabincell{c}{0.46k \\0.16k}
	& \tabincell{c}{0.48k \\0.15k}\\
\specialrule{0.05em}{1pt}{1pt}
	OntoNotes
	& \tabincell{c}{Sentences \\Entities}
	& \tabincell{c}{15.7k \\13.4k}
	& \tabincell{c}{4.3k \\6.95k}
	& \tabincell{c}{4.3k \\7.7k} \\
\specialrule{0.05em}{1pt}{1pt}
	MSRA 
	& \tabincell{c}{Sentences \\Entities} 
	& \tabincell{c}{46.4k \\74.8k}
	& \tabincell{c}{- \\-}
	& \tabincell{c}{4.4k \\6.2k} \\
\specialrule{0.1em}{1pt}{1pt}
\end{tabular}
\caption{\label{statistics} Statistics of the benchmarking datasets.}
\end{table}

We use the state of the art method, FLAT, as the baseline model. FLAT is a Chinese NER model based on Transformer and combined with lattice. Besides, we also compared the proposed method with both classic and innovative Chinese NER models.
We use the more informative `SC' as the radical feature, which comes from the online Xinhua Dictionary\footnote{\url{http://tool.httpcn.com/Zi/}.}. The pre-trained embedding of characters and words are the same as FLAT.

For hyper-parameters, we used 30 1-D convolution kernels with the size of 3 for CNN. We used the SMAC~\cite{hutter2011sequential} algorithm to search for the optimal hyper-parameters. Besides, we set a different learning rate for the training of the radical-level embedding with CNN. Readers can refer to the appendix for our hyper-parameter settings.

\subsection{Comparison with SOTA Methods}

\begin{table}[tb]
\centering\small \setlength{\tabcolsep}{7pt}
\begin{tabular}{lccc}
\specialrule{0.1em}{3pt}{3pt}
\textbf{Models} & \textbf{NE}& \textbf{NM}& \textbf{Overall}\\
\specialrule{0.1em}{3pt}{3pt}
	\citet{peng-dredze-2015-named}            & 51.96 & 61.05 & 56.05 \\
\specialrule{0.0em}{1pt}{1pt}
	\citet{peng-dredze-2016-improving}$^\ast$ & 55.28 & 62.97 & 58.99 \\
\specialrule{0.0em}{1pt}{1pt}
	\citet{he-sun-2017-f}                     & 50.60 & 59.32 & 54.82 \\
\specialrule{0.0em}{1pt}{1pt}
	\citet{10.5555/3298023.3298036}$^\ast$    & 54.50 & 62.17 & 58.23 \\
\specialrule{0.0em}{1pt}{1pt}
	\citet{cao-etal-2018-adversarial}         & 54.34 & 57.35 & 58.70 \\
\specialrule{0.05em}{1pt}{1pt}
	Lattice-LSTM                              & 53.04 & 62.25 & 58.79 \\
\specialrule{0.0em}{1pt}{1pt}
	CAN-NER                                   & 55.38 & 62.98 & 59.31 \\
\specialrule{0.0em}{1pt}{1pt}
	LR-CNN                                    & 57.14 & \textbf{66.67} & 59.92 \\
\specialrule{0.0em}{1pt}{1pt}
	LGN                                       & 55.34 & 64.98 & 60.21 \\
\specialrule{0.0em}{1pt}{1pt}
	PLT                                       & 53.55 & 64.90 & 59.76 \\
\specialrule{0.0em}{1pt}{1pt}
	SoftLexicon (LSTM)                        & 59.08 & 62.22 & 61.42 \\
\specialrule{0.05em}{1pt}{1pt}
	Baseline                          & - & - & 60.32 \\
\specialrule{0.0em}{1pt}{1pt}
	MECT                         & \textbf{61.91} & 62.51 & \textbf{63.30} \\
\specialrule{0.05em}{1pt}{1pt}
	BERT                                      & -     & -     & 68.20 \\
	BERT + MECT                          & -     & -     & \textbf{70.43} \\
\specialrule{0.1em}{1pt}{1pt}
\end{tabular}
\caption{\label{Weibo-result} Results obtained on Weibo (\%).}
\end{table}

\begin{table}[tb]
\centering\small \setlength{\tabcolsep}{8.5pt}
\begin{tabular}{lccc}
\specialrule{0.1em}{3pt}{3pt}
\textbf{Models} & \textbf{P} & \textbf{R} & \textbf{F1} \\
\specialrule{0.1em}{3pt}{3pt}
	\citet{zhang-yang2018chinese}$^\mathcal{A}$ & 93.72 & 93.44 & 93.58 \\
\specialrule{0.0em}{1pt}{1pt}
	\citet{zhang-yang2018chinese}$^\mathcal{B}$ & 94.07 & 94.42 & 94.24 \\
\specialrule{0.0em}{1pt}{1pt}
	\citet{zhang-yang2018chinese}$^\mathcal{C}$ & 93.66 & 93.31 & 93.48 \\
\specialrule{0.0em}{1pt}{1pt}
	\citet{zhang-yang2018chinese}$^\mathcal{D}$ & 94.53 & 94.29 & 94.41 \\
\specialrule{0.05em}{1pt}{1pt}
	Lattice-LSTM                      & 94.81 & 94.11 & 94.46 \\
\specialrule{0.0em}{1pt}{1pt}
	CAN-NER                           & 95.05 & 94.82 & 94.94 \\
\specialrule{0.0em}{1pt}{1pt}
	LR-CNN                            & 95.37 & 94.84 & 95.11 \\
\specialrule{0.0em}{1pt}{1pt}
	LGN                               & 95.28 & 95.46 & 95.37 \\
\specialrule{0.0em}{1pt}{1pt}
	PLT                               & 95.34 & 95.46 & 95.40 \\
\specialrule{0.0em}{1pt}{1pt}
	SoftLexicon (LSTM)                & 95.30 & \textbf{95.77} & 95.53 \\
	\quad+ bichar                     & 95.71 & \textbf{95.77} & 95.74 \\
\specialrule{0.05em}{1pt}{1pt}
	Baseline                          & - & - & 95.45 \\
\specialrule{0.0em}{1pt}{1pt}
	MECT                  & \textbf{96.40} & 95.39 & \textbf{95.89} \\
\specialrule{0.05em}{1pt}{1pt}
	BERT                              & -     & -     & 95.53 \\
	BERT + MECT                  & -     & -     & \textbf{95.98} \\
\specialrule{0.1em}{1pt}{1pt}
\end{tabular}
\caption{\label{Resume-result} Results obtained on Resume (\%). For Zhang and Yang (2018), $\mathcal{A}$ represents word-based LSTM', $\mathcal{B}$ indicates `word-based + char + bichar LSTM', $\mathcal{C}$ represents the `char-based LSTM' model, and $\mathcal{D}$ is the `char-based + bichar + softword LSTM' model.}
\end{table}

\begin{table}[t]
\centering\small \setlength{\tabcolsep}{8pt}
\begin{tabular}{lp{0.7cm}p{0.7cm}p{0.7cm}}
\specialrule{0.1em}{3pt}{3pt}
\textbf{Models} & \textbf{P} & \textbf{R} & \textbf{F1} \\
\specialrule{0.1em}{3pt}{3pt}
	\citet{Yang2018Combining}$^{\S}$              & 65.59              & 71.84              & 68.57 \\
\specialrule{0.0em}{1pt}{1pt}
	\citet{Yang2018Combining}$^{\S\ast\dagger}$ & 72.98 & \textbf{80.15} & 76.40 \\
\specialrule{0.0em}{1pt}{1pt}
	\citet{che-etal-2013-named}$^{\S\ast}$ & 77.71    & 72.51     & 75.02 \\
\specialrule{0.0em}{1pt}{1pt}
	\citet{Wang2013Effective}$^{\S\ast}$  & 76.43     & 72.32     & 74.32 \\
\specialrule{0.0em}{1pt}{1pt}
	\citet{zhang-yang2018chinese}$^{\mathcal{B}\S}$ & \textbf{78.62} & 73.13 & 75.77 \\
\specialrule{0.0em}{1pt}{1pt}
	\citet{zhang-yang2018chinese}$^{\mathcal{B}\P}$ & 73.36     & 70.12     & 71.70 \\
\specialrule{0.05em}{1pt}{1pt}
	Lattice-LSTM                          & 76.35     & 71.56     & 73.88 \\
\specialrule{0.0em}{1pt}{1pt}
	CAN-NER                               & 75.05     & 72.29     & 73.64 \\
\specialrule{0.0em}{1pt}{1pt}
	LR-CNN                                & 76.40     & 72.60     & 74.45 \\
\specialrule{0.0em}{1pt}{1pt}
	LGN                                   & 76.13     & 73.68     & 74.89 \\
\specialrule{0.0em}{1pt}{1pt}
	PLT                                   & 76.78     & 72.54     & 74.60 \\
\specialrule{0.0em}{1pt}{1pt}
	SoftLexicon (LSTM)                    & 77.28     & 74.07     & 75.64 \\
	\quad+ bichar                         & 77.13     & 75.22     & 76.16 \\
\specialrule{0.05em}{1pt}{1pt}
	Baseline                              & - & - & 76.45 \\
\specialrule{0.0em}{1pt}{1pt}
	MECT                             & 77.57     & \textbf{76.27}	    & \textbf{76.92} \\
\specialrule{0.05em}{1pt}{1pt}
	BERT                                  & -         & -         & 80.14 \\
	BERT + MECT                      & -         & -         & \textbf{82.57} \\
\specialrule{0.1em}{1pt}{1pt}
\end{tabular}
\caption{\label{Ontonotes-result} Results on Ontonotes 4.0 (\%), where `\S' denotes gold segmentation and `\P' denotes auto segmentation. }
\end{table}

In this section, we evaluate and analyze the proposed MECT method with a comparison to both the classic and state of the art methods. The experimental results are reported in Tables \ref{Weibo-result}$-$\ref{MSRA-result}\footnote{In Tables  \ref{Weibo-result}$-$\ref{MSRA-result}, `$*$' denotes the use of external labeled data for semi-supervised learning and `$\dagger$' denotes the use of discrete features.}. Each table is divided into four blocks. The first block includes classical Chinese NER methods. The second one reports the results obtained by state of the art approaches published recently. The third and fourth ones are the results obtained by the proposed MECT method as well as the baseline models.

\textbf{Weibo:} Table \ref{Weibo-result} shows the results obtained on Weibo in terms of the F1 scores of named entities (NE), nominal entities (NM), and both (Overall). From the results, we can observe that MECT achieves the state-of-the-art performance. Compared with the baseline method, MECT improves 2.98\% in terms of the F1 metric. For the NE metric, the proposed method achieves 61.91\%, beating all the other approaches.

\textbf{Resume:} The results obtained on the Resume dataset are reported in Table \ref{Resume-result}. The first block shows~\citet{zhang-yang2018chinese} comparative results on the character-level and word-level models. We can observe that the performance of incorporating word features into the character-level model is better than other models. Additionally, MECT combines lexical and radical features, and the F1 score is higher than the other models and the baseline method.

\textbf{Ontonotes 4.0:} Table  \ref{Ontonotes-result} shows the results obtained on Ontonotes 4.0. The symbol `\S' indicates gold segmentation, and the symbol `\P' denotes automated segmentation. Other models have no segmentation and use lexical matching. Compared to the baseline method, the F1 score of MECT is increased by 0.47\%. MECT also achieves a high recall rate, keeping the precision rate and recall rate relatively stable.

\textbf{MSRA:} Table \ref{MSRA-result} shows the experimental results obtained on MSRA. In the first block, the result proposed by \citet{dong2016character} is the first method using radical information in Chinese NER. From the table, we can observe that the overall performance of MECT is higher than the existing SOTA methods. Similarly, our recall rate achieves a higher performance so that the final F1 has a certain performance boosting.

\begin{table}[t]
\centering\small \setlength{\tabcolsep}{8.5pt}
\begin{tabular}{lp{0.8cm}p{0.8cm}p{0.8cm}}
\specialrule{0.1em}{3pt}{3pt}
\textbf{Models} & \textbf{P} & \textbf{R} & \textbf{F1} \\
\specialrule{0.1em}{3pt}{3pt}
	\citet{chen2006chinese}	     & 91.22          & 81.71          & 86.20 \\
\specialrule{0.0em}{1pt}{1pt}
	\citet{zhang2006word}$^\ast$ & 92.20          & 90.18          & 91.18 \\
\specialrule{0.0em}{1pt}{1pt}
	\citet{zhou2013chinese}      & 91.86          & 88.75          & 90.28 \\
\specialrule{0.0em}{1pt}{1pt}
	\citet{lu2016multi}          & -              & -              & 87.94 \\
\specialrule{0.0em}{1pt}{1pt}
	\citet{dong2016character}    & 91.28          & 90.62          & 90.95 \\
\specialrule{0.05em}{1pt}{1pt}
	Lattice-LSTM                 & 93.57          & 92.79          & 93.18 \\
\specialrule{0.0em}{1pt}{1pt}
	CAN-NER                      & 93.53          & 92.42          & 92.97 \\
\specialrule{0.0em}{1pt}{1pt}
	LR-CNN                       & 94.50          & 92.93          & 93.71 \\
\specialrule{0.0em}{1pt}{1pt}
	LGN                          & 94.19          & 92.73          & 93.46 \\
\specialrule{0.0em}{1pt}{1pt}
	PLT                          & 94.25          & 92.30          & 93.26 \\
\specialrule{0.0em}{1pt}{1pt}
	SoftLexicon (LSTM)           & 94.63          & 92.70          & 93.66 \\
	\quad+ bichar                & \textbf{94.73} & 93.40          & 94.06 \\
\specialrule{0.05em}{1pt}{1pt}
	Baseline                     & -      & -      & 94.12 \\
\specialrule{0.0em}{1pt}{1pt}
	MECT                    & 94.55          & \textbf{94.09} & \textbf{94.32} \\
\specialrule{0.05em}{1pt}{1pt}
	BERT                         & -              & -              & 94.95 \\
	BERT + MECT             & -              & -              & \textbf{96.24} \\
\specialrule{0.1em}{1pt}{1pt}
\end{tabular}
\caption{\label{MSRA-result} Results obtained on MSRA (\%).}
\end{table}

\textbf{With BERT:} Besides the single-model evaluation on the four datasets, we also evaluated the proposed method when combining with the SOTA method, BERT. The BERT model is the same as FLAT using the `BERT-wwm' released by~\citet{cui-etal-2020-revisiting}. The results are shown in the fourth block of each table. The results of BERT are taken from the FLAT paper. We can find that MECT further improves the performance of BERT significantly.

\subsection{Effectiveness of Cross-Transformer}
\begin{figure}[!t]
    \centering    
    
    \subfloat[Radical attention] 
    {
        \begin{minipage}[t]{0.2\textwidth}
            \centering          
            \includegraphics[width=1\textwidth]{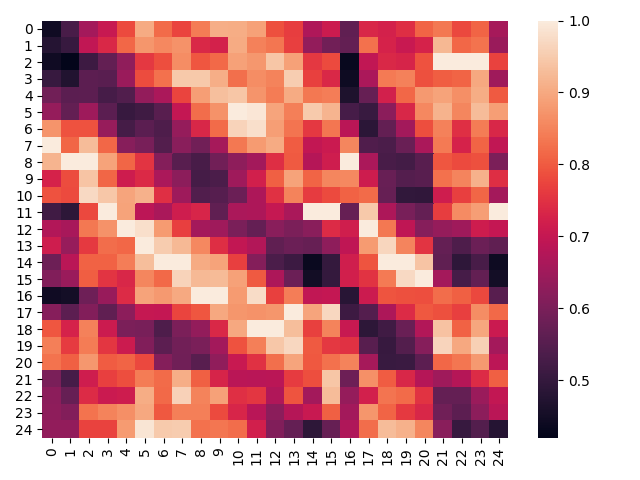}   
        \end{minipage}%
    }
    \subfloat[Lattice attention] 
    {
        \begin{minipage}[t]{0.2\textwidth}
            \centering      
            \includegraphics[width=1\textwidth]{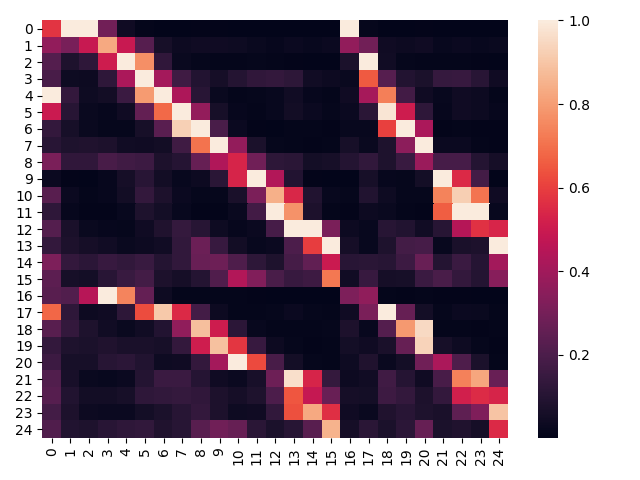}   
        \end{minipage}
    }
    
    \caption{\label{attention-visualization} Visualization of cross-attention, in which the coordinates 0-15 are used for the characters part and the coordinates 16-24 are for the words part. The two sub-figures show the radical and lattice attention scores respectively. } 
\end{figure}

There are two sub-modules in the proposed Cross-Transformer method: lattice and radical attentions.
Figure \ref{attention-visualization} includes two heatmaps for the normalization of the attention scores of the two modules. From the two figures, we can see that lattice attention pays more attention to the relationship between words and characters so that the model can obtain the position information and boundary information of words. Radical attention focuses on global information and corrects the semantic information of each character through radical features. Therefore, lattice and radical attentions provide complementary information for the performance-boosting of the proposed MECT method in Chinese NER.

\subsection{Impact of Radicals}
\begin{figure}[!t]
	\centering 
	\includegraphics[width=2in]{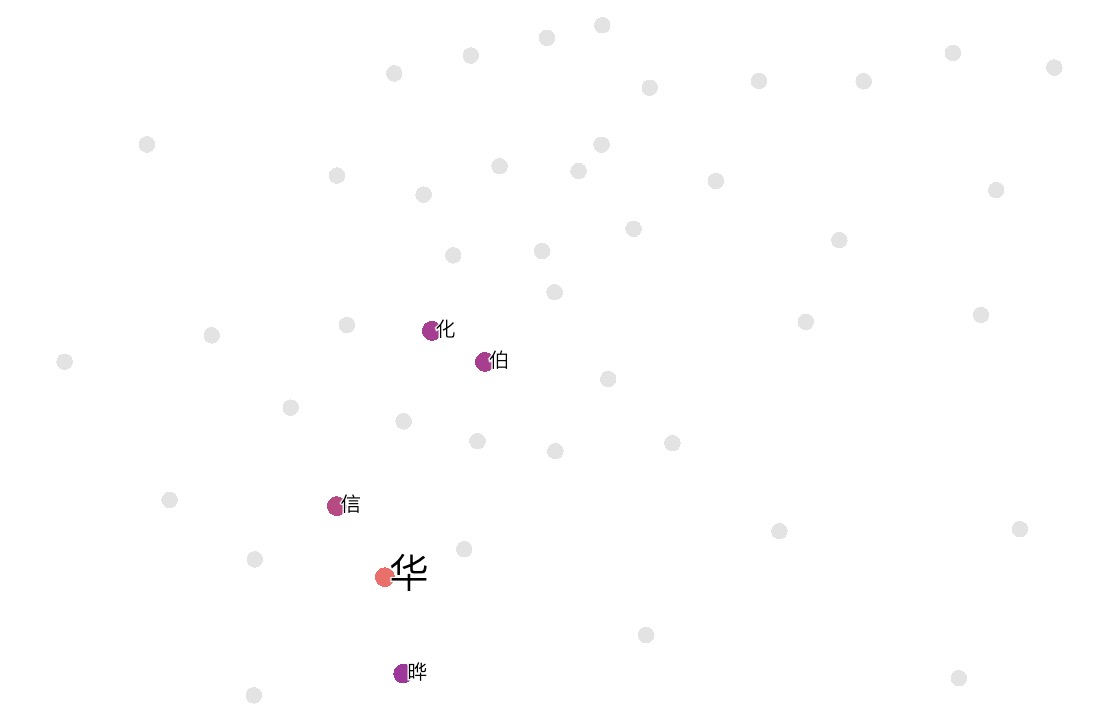} 
	\caption{\label{radical-level-embedding} Embedding visualization of the characters related to `华' in two-dimensional space. Gray dots indicate larger cosine distances.}
\end{figure}

We visualized the radical-level embedding obtained by the CNN network and found that the cosine distance of Chinese characters with the same radical or similar structure is smaller.
For example, Figure \ref{radical-level-embedding} shows part of the Chinese character embedding trained on the Resume dataset. The highlighted dots represent Chinese characters that are close to the character `华'. We can see that they have the same radicals or similar structure.
It can enhance the semantic information of Chinese characters to a certain extent.

We also examined the inference results of MECT and FLAT on \textbf{Ontonotes 4.0} and found many exciting results. For example, some words with a percentage like `百分之四十三点二 (43.2\%)' is incorrectly labelled as PER in the training dataset, which causes FLAT to mark the percentage of words with PER on the test dataset, while MECT avoids this situation. There are also some words such as `田时' and `以国' that appear in the lexicon, which was mistakenly identified as valid words by FLAT, leading to recognition errors. Our MECT addresses these issues by paying global attention to the radical information. Besides, in FLAT, some numbers and letters are incorrectly marked as PER, ORG, or others. We compared the PER label accuracy of FLAT and MECT on the test dataset. FLAT achieves 81.6\%, and MECT reaches 86.96\%, which is a very significant improvement.

\subsection{Analysis in Efficiency and Model Size}
\begin{figure} 
	\centering 
	\includegraphics[width=3in]{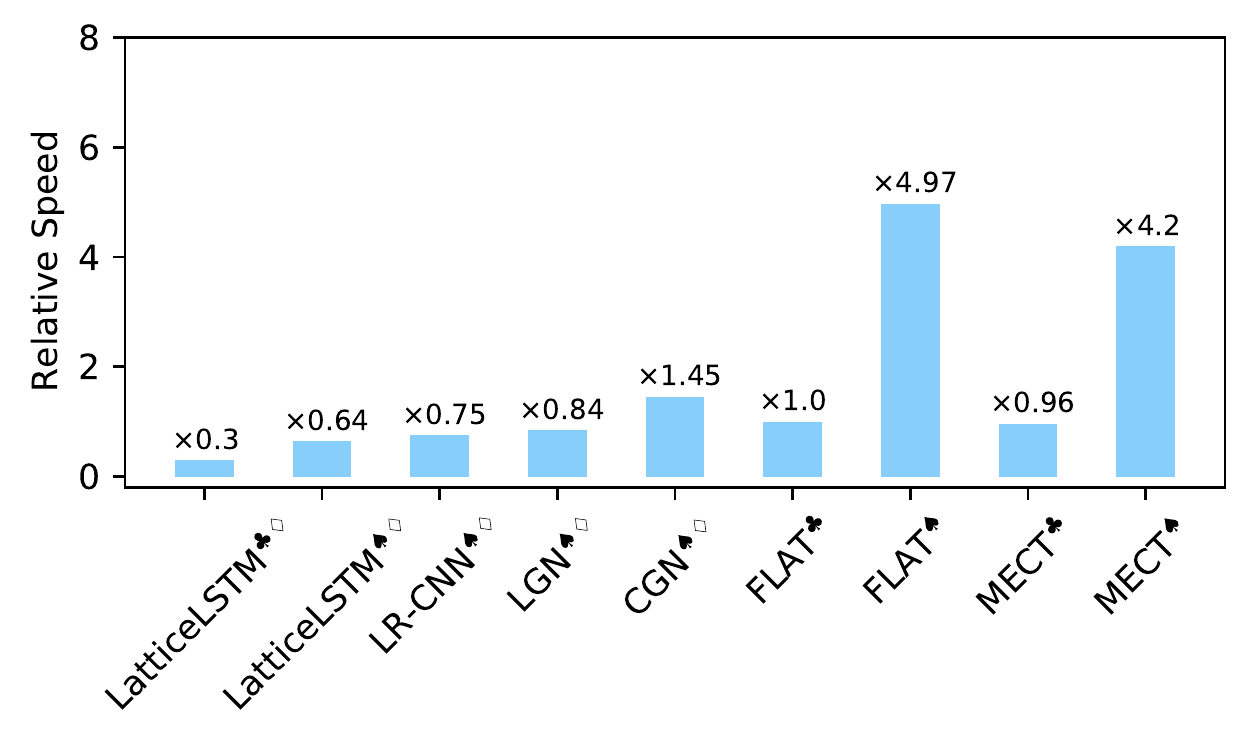} 
	\caption{\label{speed} Relative inference speed of each model based on non-parallel FLAT$^\clubsuit$. where `$\clubsuit$' represents the inference speed under non-parallel conditions, `$\spadesuit$' represents the inference speed under parallel conditions, and the value of `$\diamondsuit$' is derived from the relative speed above FLAT. }
\end{figure}

We use the same FLAT method to evaluate the parallel and non-parallel inference speed of MECT on an NVIDIA GeForce RTX 2080Ti card, using batch size = 16 and batch size = 1. We use the non-parallel version of FLAT as the standard and calculate the other models' relative inference speed. The results are shown in Figure \ref{speed}. According to the figure, even if MECT adds a Transformer encoder to FLAT, the speed is only reduced by 0.15 in terms of the parallel inference speed. Our model's speed is considerable relative to LSTM, CNN, and some graph-based network models. Because Transformer can make full use of the GPU's parallel computing power, the speed of MECT does not drop too much, but it is still faster than other models. The model's parameter is between 2 and 4 million, determined by the max sentence length in the dataset and the $d_{model}$ size in the model.

\subsection{Ablation Study}
\begin{figure}  
    \centering    
    
    \subfloat[Experiment A] 
    {
        \begin{minipage}[t]{0.24\textwidth}
            \centering          
            \includegraphics[width=1\textwidth]{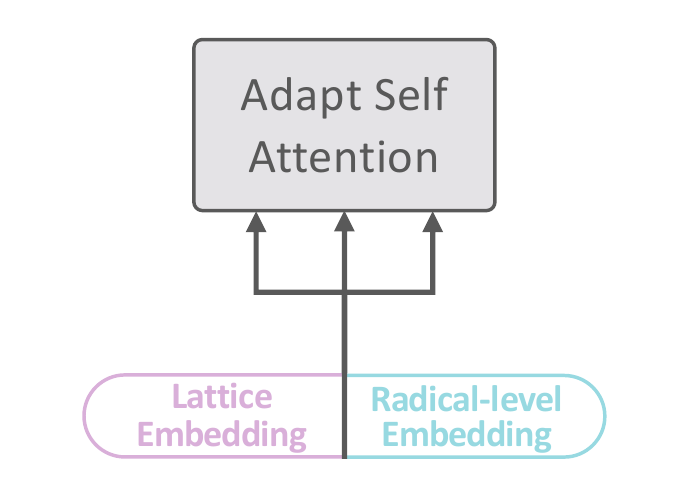}   
        \end{minipage}%
    }
    \subfloat[Experiment B] 
    {
        \begin{minipage}[t]{0.24\textwidth}
            \centering      
            \includegraphics[width=1\textwidth]{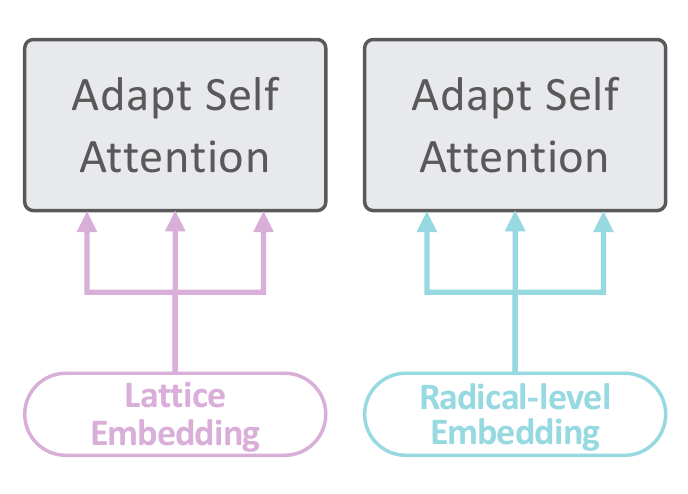}   
        \end{minipage}
    }
    \caption{\label{experiments} Two interactive attention experiment settings. } 
\end{figure}

To validate the effectiveness of the main components of the proposed method, we set up two experiments in Figure \ref{experiments}. In Experiment A, we only use a single-stream model with a modified self-attention, which is similar to the original FLAT model. The difference is that we use a randomly initialized attention matrix (Random Attention) for the attention calculation. We combine lattice embedding and radical-level embedding as the input of the model. The purpose is to verify the performance of the two-stream model relative to the single-stream model. In Experiment B, we do not exchange the query's feature vector. We replace the cross-attention with two sets of modified self-attention and follow the two modules' output with the same fusion method as MECT. The purpose of experiment B is to verify the effectiveness of MECT relative to the two-stream model without crossover. Besides, we evaluate the proposed MECT method by removing the random attention module.

Table \ref{experiment-result} shows the ablation study results. \textbf{1)} By comparing the results of Experiment A with the results of Experiment B and MECT, we can find that the two-stream model works better. The use of lattice-level and radical-level features as the two streams of the model helps the model to better understand and extract the semantic features of Chinese characters. \textbf{2)} Based on the results of Experiment B and MECT, we can see that by exchanging the two query feature vectors, the model can extract features more effectively at the lattice and radical levels. They have different attention mechanisms to obtain contextual information, resulting in global and local attention interaction. This provides better information extraction capabilities for the proposed method in a complementary way. \textbf{3)} Last, the performance of MECT drops on all the datasets by removing the random attention module (the last row). This indicates that, as an attention bias, random attention can eliminate the differences caused by different embeddings, thereby improving the model's performance further. 

\begin{table}
\centering\small \setlength{\tabcolsep}{4.5pt}
\begin{tabular}{l|cccc}
\specialrule{0.1em}{3pt}{3pt}
\textbf{Experiments} & \textbf{Weibo} & \textbf{Resume} & \textbf{OntoNotes} & \textbf{MSRA}\\
\specialrule{0.1em}{3pt}{3pt}
	\textit{Exp.} A       & 60.77 & 95.42 & 76.43 & 94.20 \\
\specialrule{0.0em}{1pt}{1pt}
	\textit{Exp.} B       & 61 & 95.54 & 76.78 & 94.18 \\
\specialrule{0.05em}{1pt}{1pt}
	MECT         & \textbf{62.69} & \textbf{95.89} & \textbf{76.92} & \textbf{94.32} \\
    \specialrule{0.0em}{1pt}{1pt}
	\quad - RA     & 61.53 & 95.31 & 76.64 & 94.25 \\
\specialrule{0.1em}{1pt}{1pt}
\end{tabular}
\caption{\label{experiment-result} The F1 scores (\%) of the four experimental methods on different datasets. RA stands for random attention. We verify all the labels (NE and NM) on Weibo.}
\end{table}

\section{Conclusion}
This paper presented a novel two-stream network, namely MECT, for Chinese NER. The proposed method uses multi-metadata embedding that fuses the information of radicals, characters and words through a Cross-Transformer network. Additionally, random attention was used for further performance boost. Experimental results obtained on four benchmarks demonstrate that the radical information of Chinese characters can effectively improve the performance for Chinese NER.

The proposed MECT method with the radical stream increases the complexity of a model. In the future, we will consider how to integrate the characters, words and radical information of Chinese characters with a more efficient way in two-stream or multi-stream networks to improve the performance of Chinese NER and extend it to other NLP tasks.

\section*{Acknowledgements}
This work was supported in part by the National Key Research and Development Program of China (2017YFC1601800), the National Natural Science Foundation of China (61876072, 61902153) and the Six Talent Peaks Project of Jiangsu Province (XYDXX-012). We also thank Xiaotong Xiang and Jun Quan for their help on editing the manuscript.

\bibliography{anthology,acl2021}
\bibliographystyle{acl_natbib}

\appendix
\section{Appendix}
\subsection{Range of Hyper-parameters}
We manually selected parameters on the two large-scale datasets, including Ontonotes 4.0 and MSRA. For the two small datasets, Weibo and Resume, we used the SMAC algorithm to search for the best hyper-parameters. The range of parameters is listed in Table \ref{hyper-parameters}.
\begin{table}[!h]
\centering
\small \setlength{\tabcolsep}{12.5pt}
\begin{tabular}{rc}
\specialrule{0.1em}{3pt}{3pt}
\textbf{Hyper-parameter} & \textbf{Range} \\
\specialrule{0.1em}{3pt}{3pt}
output\_dropout & [0.1, 0.2, 0.3]\\
\specialrule{0em}{1pt}{1pt}
lattice\_dropout & [0.1, 0.2, 0.3]\\
\specialrule{0em}{1pt}{1pt}
radical\_dropout & [0.1, 0.2, 0.3, 0.4]\\
\specialrule{0em}{1pt}{1pt}
warm\_up & [0.1, 0.2, 0.3]\\
\specialrule{0em}{1pt}{1pt}
head\_num & [8]\\
\specialrule{0em}{1pt}{1pt}
d$_{head}$ & [16, 20]\\
\specialrule{0em}{1pt}{1pt}
d$_{model}$ & [128, 160]\\
\specialrule{0em}{1pt}{1pt}
lr\ & [1e-3, 25e-4]\\
\specialrule{0em}{1pt}{1pt}
radical\_lr & [6e-4, 25e-4]\\
\specialrule{0em}{1pt}{1pt}
momentum & [0.85, 0.97]\\
\specialrule{0.1em}{3pt}{3pt}
\end{tabular}
\caption{\label{hyper-parameters} The searching range of hyper-parameters. }
\end{table}

\end{CJK}
\end{document}